\pdfoutput=1
\documentclass[sigconf]{acmart}

\usepackage{booktabs}

\AtBeginDocument{%
  \providecommand\BibTeX{{%
    \normalfont B\kern-0.5em{\scshape i\kern-0.25em b}\kern-0.8em\TeX}}}

\copyrightyear{2023} 
\acmYear{2023} 
\setcopyright{rightsretained} 
\acmConference[WWW '23 Companion]{Companion Proceedings of the ACM Web Conference 2023}{April 30-May 4, 2023}{Austin, TX, USA}
\acmBooktitle{Companion Proceedings of the ACM Web Conference 2023 (WWW '23 Companion), April 30-May 4, 2023, Austin, TX, USA}
\acmDOI{10.1145/3543873.3587320}
\acmISBN{978-1-4503-9419-2/23/04}

\newcounter{JSNumberOfComments}
\stepcounter{JSNumberOfComments}

\begin{document}

\title{Chain of Explanation: New Prompting Method to Generate Quality Natural Language Explanation for Implicit Hate Speech}
\renewcommand{\shorttitle}{Chain of Explanation}


\author{Fan Huang}
\email{huangfan@acm.org}
\affiliation{%
  \institution{Indiana University Bloomington}
  \city{Bloomington}
  \state{IN}
  \country{United States}
}

\author{Haewoon Kwak}
\email{haewoon@acm.org}
\affiliation{%
  \institution{Indiana University Bloomington}
  \city{Bloomington}
  \state{IN}
  \country{United States}
}

\author{Jisun An}
\email{jisun.an@acm.org}
\affiliation{%
  \institution{Indiana University Bloomington}
  \city{Bloomington}
  \state{IN}
  \country{United States}
}

\renewcommand{\shortauthors}{Huang et al.}


\begin{abstract}

Recent studies have exploited advanced generative language models to generate Natural Language Explanations (NLE) for why a certain text could be hateful. 
We propose the Chain of Explanation (CoE) Prompting method, using the heuristic words and target group, to generate high-quality NLE for implicit hate speech. 
We improved the BLUE score from 44.0 to 62.3 for NLE generation by providing accurate target information. We then evaluate the quality of generated NLE using various automatic metrics and human annotations of informativeness and clarity scores.

\end{abstract}




\maketitle

\section{Introduction}

\vspace{0.5mm}
\textbf{Warning:} \textit{This paper contains offensive content and may be upsetting.}
\vspace{0.2mm}

Tremendous hateful speeches are created and spread every second online,
which can lead to various social problems~\cite{hine2017kek}.
Natural language processing has shown to be a powerful tool to accurately and efficiently detect hateful speech on online social platforms~\cite{salminen2018anatomy}. 

When a text \emph{explicitly} contains hateful speech and obvious discrimination words, feature attribution approaches, such as LIME~\cite{LIME} or SHAP~\cite{NIPS2017_SHAP}, can provide reliable information on why the text should be classified as harmful by highlighting specific words.

To obtain plausible and faithful explanations for implicit hate speech, various natural language explanation (NLE) generation methods have been proposed.
However, most previous studies have focused on autoregressive generative language models (GLMs) to generate NLE for hate speech without prompting methods~\cite{elsherief-etal-2021-latent}. 
The potential of sequence-to-sequence (Seq2Seq) models and prompting methods has not been fully explored~\cite{ding2021openprompt}. 
Moreover, traditional evaluation metrics, such as BLEU~\cite{BLEU} and Rouge~\cite{lin-2004-rouge}, applied in NLE generation for hate speech, may also not be able to comprehensively capture the quality of the generated explanations because they heavily rely on the word-level overlaps~\cite{clinciu-etal-2021-study}. 
To fill those gaps, we propose a Chain of Explanations (CoE) prompt method to generate high-quality NLE distinguishing the implicit hate speech from non-hateful tweets. We then benchmark the mainstream Natural Language Generation models through comprehensive auto-evaluation metrics and human evaluation.

\section{Related Work}

\noindent \textbf{Hate Speech Detection and Explanation.} Due to its societal importance, researchers have actively studied and proposed various models to detect hate speech~\cite{elsherief-etal-2021-latent, IsChatGPT2023}.
Providing explanations to the AI system users would help improve user experience and system efficacy~\cite{epstein2022explanations}. 
Furthermore, providing implied meanings of the text before it is posted has proven to help prevent the potentially harmful posts~\cite{elsherief-etal-2021-latent}. 
For those explicit hate speech, highlighting words or phrases, which can be done via feature attribution-based explainable techniques like LIME~\cite{LIME} or SHAP~\cite{NIPS2017_SHAP}, post-hoc explanations techniques that rely on input perturbations, could be an effective way to explain. 
Thus, more recent works were proposed to apply the Generative Pre-trained models to create hate explanations~\cite{elsherief-etal-2021-latent}.


\noindent \textbf{Prompt Learning.} In recent years, prompt learning has become a widely-accepted paradigm for pre-trained language models~\cite{ding2021openprompt}. 
The prompt learning could mine knowledge from the pre-trained models in multiple manners through manually selected prompt designs~\cite{gao2020making}. 
With the help of prompt learning, the text generation task could be fostered through a prompt with task-specific information.
Still, challenges for prompt learning are: (1) the well-performed prompts are highly task-specific, and (2) the existing well-performed prompt may not be suitable for all data instances when facing different data instances~\cite{gao2020making}. 
While promising, it is under-explored whether prompt learning can generate quality explanations for implicit hate speech.

\section{Chain of Explanation}

\subsection{Baseline Generation Method}

Following the text generation task formulation used in \cite{elsherief-etal-2021-latent}, we construct our baseline task that only generates NLE for a given text. During the training process, the generation model uses  the following token sequence as the input:
\begin{multline*}
x=\{[S T R], t_{1}, t_{2}, \ldots, t_{n},[S E P],
 t_{E 1}, t_{E 2}, \ldots, t_{E m}, [E N D]\}
\end{multline*}
where [STR] is the start token,  [SEP] is the separate token, and  [END] is the end token. Inside,  $t_{1}, ..., t_{n}$ represents an input tweet, while $t_{E1}, ..., t_{Em}$ represents the NLE. 

\subsection{Chain of Explanation Prompt}

Inspired by \cite{wei2022chain} that proves a  chain of thought prompting is effective for various tasks involving a complex reasoning process, we propose the Chain of Explanation prompting method for implicit hate speech explanation generation. Our prompt is based on the following guidelines: 
(1) heuristic words to inform the expected information in the prompting structure,
(2) demonstration of the hateful intention of the given text, and 
(3) demonstration of the target group of the hateful intention. 

For the prompt design, the input token sequence is as follows:
\begin{equation*}
\begin{split}
x & =\{[S T R], H_{text}, t_{1}, t_{2}, \ldots, t_{n},[S E P], \\ &
\quad D_{hate}, [S E P], \\ & \quad D_{target}, [S E P], \\ & \quad H_{NLE},
t_{E 1}, t_{E 2}, \ldots, t_{E m}, [E N D]\}
\end{split}
\end{equation*}
where \(H_{text}\) stands for the heuristic words for a given text, which is the tokenized sentence of ``Given Text: '';
\(D_{hate}\) stands for the demonstration information for hateful intention, which is the tokenized sentence of ``Is the text hateful? Yes'';
 \(D_{target}\) stands for the demonstration information for target group of hateful intention, which is the tokenized sentence of ``The target group is: \{target\}''; and 
\(H_{NLE}\) stands for the heuristic words for NLE, which is the tokenized sentence of ``It is hateful because: ''.

\section{NLE Generation}

\subsection{Dataset}

To train and test the NLE generation for hate speech, we use the \textit{LatentHatred} dataset~\cite{elsherief-etal-2021-latent}.
The LatentHatred dataset includes 6,358 \textit{implicit} hateful tweets, and each tweet is annotated with its hate category, target group (a particular group of people (e.g., Asian or women) targeted by hate speech), and implied statement (the implications of the implicit hateful intention).

\subsection{Models}

The GPT and GPT-2 models are widely used in the generation tasks for complex reasoning.
As the accessibility to the GPT-3 model is limited compared to its preceding models, we use GPT-NEO~\cite{gpt-neo} and OPT~\cite{zhang2022opt} model instead. 
The Seq2Seq models, generating the sequence conditioned on the input sequence, could also generate the NLE based on the provided information. The most widely used Seq2Seq models are T5~\cite{2020t5} and BART~\cite{lewis-etal-2020-bart}. 
As for the generation settings, we choose greedy decoding for all the above models.
For our experiment, we use the basic version of those models: ``gpt2'', ``gpt-neo-125m'', ``opt-125m'', ``bart-base'', and ``t5-base''.

\subsection{NLE Evaluation Metrics}

The BLEU~\cite{BLEU} and ROUGE~\cite{lin-2004-rouge}  metrics have been commonly used to measure the quality of generated texts by using the word overlaps. 
The Meteor~\cite{banerjee-lavie-2005-meteor} metric calculates the score based on the harmonic mean of unigram recall and precision values considering synonym and stemming. 
The NIST metric evaluates the informativeness of the n-grams~\cite{doddington2002automatic}. 
The SARI metric measures the goodness of words by comparing generated NLE with the ground truth NLE~\cite{xu2016optimizing}. 
With the help of Transformers, semantic-based metrics, such as  BERTScore~\cite{bert-score} and BLEURT~\cite{sellam2020bleurt},  make use of the word embedding similarity.
The NUBIA metric, known as the most advanced metric, reflects how interchangeable the sentences are using various neural models ~\cite{kane2020nubia}.  All metrics ranges from 0 to 100 except NIST (0 to 300) and BERTScore (-100 to 100). For all scores, the higher, the better.




\section{Evaluation}

Table~\ref{Example_reviews} shows an example of generated explanations given a implicit hateful tweet along with ground truth explanation.  
We see that the results of our proposed models (GPT-2,CoE and BART,CoE) are comparable with the baseline model (GPT-2,base).

\begin{table}[h!]
\begin{tabular}{ll}
\toprule
 & \textbf{Generated Explanation} \\
 \midrule
 Ori. Tweet & illegal immigrants are not americans \\
  & dems have no right to an opinion depo... \\
\midrule
 Ground Truth & Illegal immigrants should get deported \\
 GPT-2 (base) & Immigrants are not citizens of the US \\
 GPT-2 (CoE) & Immigrants are not Americans \\
 BART (CoE) & Immigrants are taking over society \\
\bottomrule
\end{tabular}
\caption{\label{Example_reviews}
An example of generated explanations.}
\end{table}

\begin{table*}[h!]
\centering
\begin{tabular}{clccccccccccc}
\toprule
& Models & B1 & B2 & M & NI & R1 & R2 & R-L & S & BS & BL & NU\\
\toprule
\cite{elsherief-etal-2021-latent} & GPT-2* & 42.3 & - & - & - & - & - & 32.7 & - & - & - & - \\
\midrule
Baseline & GPT-2  &  44.0 & 20.5  & 30.4  & 63.9  & 33.2  & 16.8  & 33.0  & 39.7  & 57.0  &  45.1 & 25.3  \\
& GPT-Neo  &  \textbf{44.2} & 20.4  & 30.0  & 64.1  & 34.1  &  17.0 & \textbf{33.7}  & 39.9  & 57.7  & 44.1  & 24.0  \\
& OPT  & \textbf{44.2}  & 20.8  &  31.3 & 66.3  & \textbf{34.3}  &  17.7 & 34.0  & 39.9  & 57.9  & 45.5  & 26.8  \\
& BART  & 43.2  & \textbf{21.4}  & \textbf{31.6}   &  \textbf{68.3} & 33.3  & \textbf{17.9}  & 33.1  & \textbf{40.1}  & \textbf{58.3}  & \textbf{46.3}  & \textbf{29.2}  \\
& T5  & 41.6  & 18.9  & 29.3  & 62.8  & 31.8  & 15.8  & 31.5  & 39.5  & 55.6  & 43.6  &  23.2 \\
\midrule
CoE & GPT-2 & 59.0 & 39.8  &  46.9  & 91.8  & 51.1  & 34.9  & 50.9  & 45.2  & 68.6  & 53.9  &  38.1  \\
& GPT-Neo  &  56.8 &  37.7 & 45.8  & 90.2  & 48.8  & 32.5  & 48.6  & 42.9  &  64.6 & 50.8  & 33.5  \\
& OPT  & 59.5  &  40.8 & \textbf{48.3} & \textbf{94.3}  & 51.7  &  35.7 & 51.6  & 43.9  & 69.1  & 54.0  & 38.1  \\
& BART  & \textbf{61.8}  & \textbf{42.7}  & 47.6  & 93.8  &  \textbf{52.7} & \textbf{36.4}  & \textbf{52.6}  & \textbf{45.6}  & \textbf{70.6}  & \textbf{54.9}  & \textbf{41.9}  \\
& T5  & 58.6  & 39.4  &  46.9  &  89.9  & 50.7  & 34.6 &  50.6 & 45.0  & 68.2  & 53.4  & 37.6  \\
\bottomrule
\end{tabular}
\caption{\label{baseline}
Evaluation of the generated explanation in comparison with baseline models. 
The first row shows the experiment result reported in \cite{elsherief-etal-2021-latent}.
Different columns represent the different auto-evaluation metrics: BLEU unigram (B1), BLEU bigram (B2), Meteor (M), NIST(NI), f1-score of Rouge-1 (R1), Rouge-2 (R2), and Rouge-L (R-L), SARI (S), f1-score of the BERTScore (BS), BLEURT (BL), and NUBIA (NU).} 
\end{table*}

We evaluate our results in comparison with the reported baseline~\cite{elsherief-etal-2021-latent} and our replication of the baseline results. 
\citet{elsherief-etal-2021-latent} use the GPT-2 model to generate both the target group and implied statement at once without a prompt method. 
We follow a similar training setting and tuning process of hyper-parameters in \cite{elsherief-etal-2021-latent}. We fine-tune for \(e \in \{1,2,3,4,5\}\) with the batch size per device of 2 and learning rate of \(5 \times 10^{-5}\), also with 100 steps of linear warm up. Our split portion is 75:12.5:12.5 for training, testing, and validation.

Table~\ref{baseline} shows the results across various automatic metrics explained in $\S$4.3. 
Our replications show comparable results with the reported baselines. 
For the baseline method, we find the OPT model performs the best among the autoregressive models, and the BART model performs the best among Seq2Seq models. Generally, BART outperforms other models.

At the row of the chain of explanation (CoE) in the table, we can see the significant improvement across all automatic evaluation metrics. 
The automatic evaluation metric scores improve significantly from 44.2 to 61.8 for BLUE-1 and from 33.7 to 52.6 for Rouge-L. 
Similarly, the OPT remains the best model among autoregressive models, and the BART remains the best among Seq2Seq models. Between them, BART outperforms in most of the metrics. 

\begin{table*}[h!]
\centering
\begin{tabular}{lccccccccccc}
\hline
Models & B1 & B2 & M & NI & R1 & R2 & R-L & S & BS & BL & NU\\
\hline
BART  & 61.8  & 42.7  & 47.6  & 93.8  &  52.7& 36.4  & 52.6  & 45.6 & 70.6  & 54.9  & 41.9  \\
Heuristic*  & 59.6  & 39.9  & 46.6  & 93.0  & 51.5  & 34.5  & 51.4  & 45.1   &  70.1 &  55.5 & 41.4 \\
Hate Label*  &  61.9 &  42.2 & 47.3  & 92.3  &  52.4 & 36.2  & 52.4  & 45.6  & 70.4  & 54.6  & 40.9  \\
Target Group*  &  45.5 &  23.9 & 34.0  & 72.2  &  36.1 & 20.1  & 35.8  & 41.2  & 59.8  & 47.3  &  30.5 \\
\hline
\end{tabular}
\caption{\label{ablation_autoregressive}
The ablation test results of the chain of explanation prompt design for the Seq2Seq model BART, where the first row records the scores from the best model in Table~\ref{baseline}. 
The next three rows with an asterisk show the result by removing that feature (e.g., Target Group* stands for the ablation test that removes target group information). }
\end{table*}

\noindent \textbf{Ablation Study}
We further study the importance of each part of our CoE prompt design by conducting an ablation study. 
Table~\ref{ablation_autoregressive} present the ablation test results for BART, the best performing Seq2Seq model, inspecting three variations of the chain of explanation prompting design. 
The ablation design is to remove the heuristic words and re-run the test to see if there would be a difference. 
We conduct a similar ablation study for OPT, the autoregressive model, and find consistent results. 

(1) \textbf{The heuristic words.} We find that the BLEU-1 score drops by 0.6 when removing the heuristic works. One explanation for why the chain of explanation prompting method performs well is that the heuristic words added to the text contributed to guiding the PLMs to generate the explanation with the clear purpose of answering why the given tweets would be considered hateful. 

(2) \textbf{The hate label demonstrations.} The BLEU-1 score drops by 1.4 for the BART model. Another possible explanation is that the hate label helps the model to confirm the hateful nature of the given text and thus gives a hint to the model that the given text should not be understood as the non-hateful text so that the model would not generate the meaningless NLE.

(3) \textbf{The target group demonstrations.} We observe the target group is the most important information in generating high quality hate explanation. Removing the target group results in dropping 15.5 of the BLEU-1 score. 


\noindent \textbf{Human Evaluation - Informativeness and Clarity} Human perceptions have always been considered one of the most important evaluation criteria for Natural Language Generation tasks~\cite{evans2018learning,gkatzia-mahamood-2015-snapshot}. 
Thus, we investigate the quality of the generated NLE from the perspective of human perception by using two metrics\textemdash\textit{Informativeness} and \textit{Clarity}. 
Informativeness~\cite{duvsek2020evaluating} considers how relevant the information in NLE explains why the tweet should be perceived as hateful by human readers (e.g., 1 = Not Informative and 7 = Very Informative). 
Clarity~\cite{clarity_NLG} measures how clear the meaning of the NLE is expressed (e.g., 1 = Unclear and 7 = Very Clear). 

For human annotation, we first tried Amazon Mechanical Turk (AMT). However, we found that the raters do not agree with each other---the inter-rater reliability (Krippendorff's Alpha value) was 0.14. 
Thus, we hired three experienced Research Assistants to annotate our data. 
For 100 randomly selected NLE samples, we collect at least three annotations for each NLE. We then remove the instances that are too hard to reach a consensus based on the same rule in \cite{clinciu-etal-2021-study}. 
This results in annotations for 68 NLE samples.
The inter-rater reliability score is 0.35, indicating that the raters are in relatively fair agreement. 
For the ground truth NLE, the average Informativeness is 5.20 (95\% CI: 4.95---5.45), and the average Clarity is 4.53 (95\% CI: 4.30---4.77).
Our best generated NLE shows slightly lower performance on both metrics---its average Informativeness is 4.48 (95\% CI: 3.95---5.00), and the average Clarity is 4.34 (95\% CI: 3.95---4.72). 
Still, these values are comparable with the results of existing work~\cite{clinciu-etal-2021-study}. 
In interpreting Bayesian Network graphical representations, which could be a less subjective task than hate explanation, the human written explanations achieved, on average, 4.66 (95\% CI: 4.44---4.88) and  4.65 (95\% CI: 4.45---4.86) for Informativeness and Clarity, respectively. 




We further analyze how those two human-annotated metrics are correlated with automatic evaluation metrics to show the gap between them. 
For comparison, we use the median scores of the human annotations, as proposed in \cite{clinciu-etal-2021-study}, and the scores of the automatic metrics by our best model (i.e., BART).
Table~\ref{significance} presents the Spearman correlation coefficients between each of all automatic metrics and the two human annotated scores.

First, the BLEURT, BERTScore, and NUBIA metrics correlate better with the two human evaluation metrics. The BLEURT metric shows the highest correlations. Second, the SARI metric shows the lowest correlations with both Informativeness and Clarity. 
Lastly, for the widely used BLEU-1 and ROUGE-L metrics, the correlation with Informativeness is significantly higher than Clarity, which is not aligned with \cite{clinciu-etal-2021-study}. Understanding the origin of the differences would be a good future research direction. 

\begin{table}
\centering
\begin{tabular}{lccc}
\hline
Metrics & Informativeness & Clarity\\
\hline
BLEU-1 (B1) & 0.31* & 0.15 \\
BLEU-2 (B2) & 0.25 & 0.12 \\
Meteor (M) & 0.29* & 0.17 \\
NIST (NI) & 0.27* & 0.22 \\
Rouge-1 (R1) & 0.32* & 0.18 \\
Rouge-2 (R2) & 0.27 & 0.14 \\
Rouge-L (R-L) & 0.32* & 0.18 \\
SARI (S) & 0.16 & 0.051 \\
\hline
BERTScore (BS) & \textbf{0.34*} & \textbf{0.26*} \\
BLEURT (BL) & \textbf{0.5*} & \textbf{0.41*} \\
\hline
NUBIA (NU) & \textbf{0.48*} & \textbf{0.35*} \\
\hline
\end{tabular}
\caption{\label{significance}
Spearman Rank Correlation between automatic and human evaluation metrics. * indicates p $<$ 0.05.}
\end{table}

\section{Discussion and Conclusion}

In our work, we proposed the Chain of Explanation prompt design to generate high-quality Natural Language Explanations for implicit online hate speech. The performance is evaluated comprehensively through various auto-evaluation metrics and human evaluation.

As the prompting method aims to generate Natural Language Explanations to illustrate why the given text is toxic or hateful,  model's output thus could contain hateful expressions, which may produce additional stress to the end users. 
The PLMs would also learn the implicit hateful or conspiracy-based logic and expressions~\cite{levy2021investigating}, making the generation results  less accountable. The potential solution is to apply a shepherding system to filter out hateful statements or transfer them to non-hateful expressions.

\section{Ethical Considerations}

The Institutional Review Board at Singapore Management University has approved this study 
 (Approval No.: IRB-22-076-A043(622)). 
For the annotation process, we included a warning in the instructions informing that the content might be offensive or upsetting. Annotators were also encouraged to stop the labeling process at any time if they felt overwhelmed or unwell. 







\bibliographystyle{ACM-Reference-Format}
\bibliography{references}

\end{document}